\documentclass[10pt,twocolumn,letterpaper]{article}

\usepackage{cvpr}              
\usepackage[accsupp]{axessibility}
\usepackage{graphicx}
\usepackage{amsmath}
\usepackage{amssymb}
\usepackage{booktabs}
\usepackage{amsmath}
\usepackage{multirow}
\usepackage{graphicx}
\usepackage{amsfonts}
\usepackage{amsthm}
\usepackage{multirow}
\usepackage{dsfont}
\newtheorem{theorem}{Theorem}[section]

\theoremstyle{definition}

\newtheorem{proposition}{Proposition}[section]

\newtheorem{lemma}{Lemma}[section]

\usepackage[pagebackref,breaklinks,colorlinks]{hyperref}

\usepackage[capitalize]{cleveref}
\crefname{section}{Sec.}{Secs.}
\Crefname{section}{Section}{Sections}
\Crefname{table}{Table}{Tables}
\crefname{table}{Tab.}{Tabs.}

\def\cvprPaperID{4945} 
\def\confName{CVPR}
\def\confYear{2022}

\begin{document}

\title{Improving Subgraph Recognition with Variational Graph Information Bottleneck}

\author{Junchi Yu$^{1,2}$,
Jie Cao$^{1}$,
Ran He$^{1,2,3}$\thanks{Corresponding Author}\\
\and
$^1$NLPR\&CRIPAC Institute of Automation, Chinese Academy of Sciences, China\\
$^2$University of Chinese Academy of Sciences, China\\
$^3$Center for Excellence in Brain Science and Intelligence Technology, CAS, China\\
{\tt\small yujunchi2019@ia.ac.cn, jie.cao@cripac.ia.ac.cn, rhe@nlpr.ia.ac.cn}
}
\maketitle

\begin{abstract}
Subgraph recognition aims at discovering a compressed substructure of a graph that is most informative to the graph property. 
It can be formulated by optimizing Graph Information Bottleneck (GIB) with a mutual information estimator. 
However, GIB suffers from training instability and degenerated results due to its intrinsic optimization process.
To tackle these issues, we reformulate the subgraph recognition problem into two steps: graph perturbation and subgraph selection, leading to a novel Variational Graph Information Bottleneck (VGIB) framework. VGIB first employs the noise injection to modulate the information flow from the input graph to the perturbed graph. Then, the perturbed graph is encouraged to be informative to the graph property. VGIB further obtains the desired subgraph by filtering out the noise in the perturbed graph. With the customized noise prior for each input, the VGIB objective is endowed with a tractable variational upper bound, leading to a superior empirical performance as well as theoretical properties. 
Extensive experiments on graph interpretation, explainability of Graph Neural Networks, and graph classification show that VGIB finds better subgraphs than existing methods \footnote{Code is avaliable on \url{https://github.com/Samyu0304/VGIB}}.
\end{abstract}

\section{Introduction}
\begin{figure}[t]
\begin{center}
\centerline{\includegraphics[width=1.0\columnwidth]{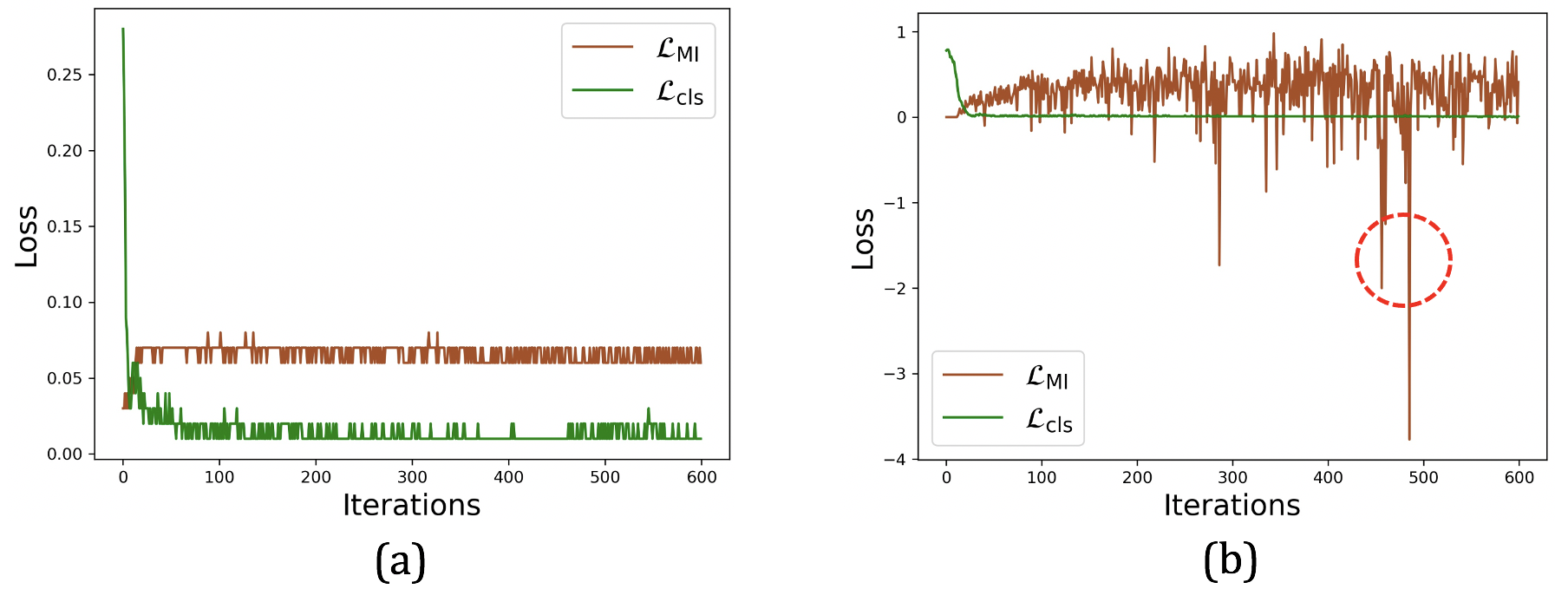}}
\end{center}
\vspace{-1.2cm}
\caption{Training dynamics of VGIB and GIB. $\mathcal{L}_{\mathrm{cls}}$ and $\mathcal{L}_{\mathrm{MI}}$ refer to the prediction and compression term of two methods.  (a). VGIB converges fast and stable. (b). GIB suffers from an unstable training process and inaccurate estimation of mutual information in the red circle (mutual information is non-negative).}
\label{gib_dy}
\vspace{-0.5cm}
\end{figure}

Graph classification, which aims to identify the labels of graph-structured data, has attracted much attention in diverse fields such as biochemistry \cite{mmpn,jin2018junction,jin2020multi,rong2020self}, social network analysis \cite{gcn,conf/nips/HamiltonYL17,velickovic2017graph}, and computer vision \cite{chen2020hierarchical,li2019deepgcns,lei2020seggcn,lin2020convolution}. 
Recently, there has been a surge of interest in its reverse problem. 
That is, to recognize a compressed subgraph of the input, which is most predictive to the graph label \cite{yu2020graph}.
Such a subgraph enjoys superior property for predictive performance since it drops noisy and redundant information and only preserves label-relevant information \cite{yu2021recognizing, wu2020graph}. Meanwhile, the produced subgraph serves as an intrinsic explanation to the prediction of the graph model \cite{gnnexplainer}.
Hence, recognizing a compressed yet informative subgraph, namely the subgraph recognition, is the fundamental problem of many tasks.
For example, biochemists are interested in discovering the substructure of the molecule which most affects the molecule properties \cite{yu2021recognizing, jin2020multi}.
In the explainability of Graph Neural Networks (GNNs), it is vital to generate the explanatory subgraph of the input, which faithfully interprets the predicted results \cite{gnnexplainer, luo2020parameterized}.
In graph classification, the significant substructures, such as nodes, edges, and subgraphs, are highlighted to improve the predictive performances \cite{sun2019infograph,alsentzer2020subgraph,li2020graph}. The subgraph recognition problem is first studied in a unified view under the Graph Information Bottleneck (GIB) framework \cite{yu2020graph}. It employs the Shannon mutual information to quantify the compressed and informative nature of the subgraph distribution.
Although GIB permits theoretical analysis of the subgraph recognition problem, its optimization process is inefficient and unstable due to mutual information estimation, which is shown in Fig.~\ref{gib_dy}. Meanwhile, the inaccurate estimated value leads to degenerated performance on subgraph recognition.
These issues motivate us to advance the existing framework for improved subgraph recognition.

In this work, we address all the above issues with the proposed Variational Graph Information Bottleneck (VGIB).
VGIB reformulates the subgraph recognition problem into two steps: graph perturbation and subgraph selection.
During graph perturbation, VGIB employs a noise injection method to selectively inject noises into the input graph to obtain the perturbed graph. 
The intuition is that the noise injection naturally modulates the information flow from the input graph to the perturbed graph.
Specifically, the more injected noise leads to more significant information distortion in the perturbed graph, which in analogy to the compressed nature of the subgraph.
Hence, VGIB approaches the compression condition in GIB with the total amount of injected noise in the perturbed graph, leading to a tractable \textit{variational} upper bound.
Meanwhile, VGIB encourages the perturbed graph to be informative of the graph property, which indicates less injected noise.
This trade-off condition guides the noise injection module to only inject noise into the insignificant substructure while preserving the predictive portion of the input.
However, the design of noise injection is non-trivial. First, the action space of noise injection is discrete, leading to the difficulty of optimization with gradient methods.
Hence, we employ the Gumbel-Softmax reparametrization for noise injection.
Secondly, injecting random noise will break the semantic information of the input and lead to the difficulty of noise quantification.
To this end, we customize the Gaussian prior for each input graph.
With the above configurations, the VGIB objective enables a tractable variational upper bound and is efficient to optimize the gradient-based method. 
After training VGIB with graph perturbation, only the insignificant substructure of the graph is perturbed and the informative subgraph is well-preserved.
Thus, in the subgraph selection step, one can obtain the found subgraph by dropping the injected noise.

We evaluate the proposed VGIB framework on various tasks, including explainability of GNNs, graph interpretation, and graph classification. 
The experimental results show that VGIB enjoys significant efficiency in optimization and outperforms the baseline methods with better found subgraphs.

\section{Related Work}


\textbf{Information Bottleneck.} The information bottleneck (IB) principle attempts to juice out a compressed but predictive code of the input signal \cite{tishby2000information}. Alemi et al.~\cite{alemi2016deep} first empowers deep learning with a variational information bottleneck (VIB). Currently, the applications of IB and VIB in deep learning are mainly attributed to representation learning and feature selection. In the representation learning scenario, researchers employ a deterministic or stochastic encoder to learn a compressed yet meaningful representation of the input data, to facilitate various downstream tasks, such as computer vision \cite{rl_ib1,conf/iccv/LuoLGYY19}, reinforcement learning \cite{rl_ib2,rl_ib3}, natural language processing \cite{nlp_ib1}, speech and acoustics \cite{speech_ib1}, and node representation learning \cite{wu2020graph}. For the feature selection, IB is used to select a subset of input features such as pixels in images or dimensions in vectors, which are maximally predictive to the label of input data \cite{achille2018information,schulz2020restricting,kim2021drop}. \cite{achille2018information,schulz2020restricting} inject noises into the intermediate representations of a pretrained network and select the areas with maximal information per dimension. \cite{kim2021drop} learns the drop rates for each dimension of the vector-structured features. Unlike the prior work on the regular data, Yu et al.~\cite{yu2020graph} first recognize a predictive yet compressed subgraph from the irregular graph input and thus facilitates various graph-level tasks.

\textbf{Graph Classification.} The goal of graph classification is to infer the label or property of an input graph.  Recently, there is a surge of interest in applying the Graph Neural Network (GNN) for graph classification \cite{sortpool,asapool}. It first aggregates the messages in the neighborhoods for node representations, which are pooled for the graph representations for prediction by a readout function. The typical implementations of readout are mean and sum functions \cite{gcn,velickovic2017graph,huang2018adaptive,Xu:2019ty}. Besides, it is popular to leverage the hierarchical and more complex information in the graph, which leads to the graph-pooling methods \cite{sortpool,hierG,asapool,specpool,diffpool}. These methods generally leverage all the information in graphs for prediction, neglecting the importance of the informative substructure. Hence, subgraph recognition can enhance graph classification with the label-relevant information in the graphs \cite{yu2020graph}.



\textbf{Subgraph Discovery.} Subgraph discovery in the literature of traditional data mining refers to discovering subgraphs with specific topology \cite{densesub,conf/kdd/GionisT15,gspan}. It is similar to the data generation task \cite{yu2019pose,sun2021multi}. Recently, there is a trend to leverage the importance of subgraphs in graph learning. At node-level tasks, researchers focus on passing the message of a neighborhood subgraph to the central node \cite{chen2018fastgcn,conf/nips/HamiltonYL17,huang2018adaptive}. NeuralSparse \cite{zheng2020robust} chooses the most relevant K neighborhoods of a central node for robust node classification. At the graph level, it is popular to discover the information in subgraphs for learning graph representations. Infograph \cite{sun2019infograph} maximize the mutual information between representations of graphs and the corresponding local patches. 
Another direction closely related to subgraph recognition is to interpret a pretrained GCN with interpretable subgraphs. GNNExplainer \cite{gnnexplainer} discover the neighborhood subgraph, which maximally affects the prediction of the central node. SubgraphX \cite{yuan2021explainability} explains the prediction of GCN with a subgraph found by Monte Carlo Tree Search. However, these methods only explore the subgraph recognition problem with specific tasks, and thus lack an unified view of the subgraph recognition problem.

\section{Notations and Backgrounds}
In this section, we introduce our notations and preliminaries. 
Let $G=\{A,X\}\in \mathbb{G}$ be a graph with $n$ nodes, with $A \in \mathbb{R}^{n\times n}$ and $X \in \mathbb{R}^{n\times d}$ being its adjacent matrix and node feature matrix. 
We denote $\{(G_{1},Y_{1}),(G_{2},Y_{2}),\cdots,(G_{n},Y_{n})\}$ as the set of $n$ graphs with its corresponding categorical labels or real-value properties. 
We use $G_{sub}$ to denote the subgraph in $G$.
We denote $I(X,Y)$ as the mutual information between the random variables $X$ and $Y$, which takes the form:
\begin{equation}
\begin{aligned}
\notag I(X,Y) =\int_{X}\int_{Y}p(x,y)\log{\frac{p(x,y)}{p(x)p(y)}}\mathrm{d}x\mathrm{d}y
\end{aligned}
\end{equation}

\subsection{Graph Information Bottleneck}
Given the input data $X$ and its label $Y$, the information bottleneck (IB) \cite{tishby2000information} principle learns the \emph{minimal sufficient} representation $Z$ by optimizing the IB objective:
$\min_{Z}{-I(Z,Y) + \beta I(Z,X)}.$
Here $\beta$ is the Lagrangian parameter to balance the two terms. Inspired by it, Yu et al. propose Graph Information Bottleneck (GIB) principle to recognize an informative yet compressed subgraph from the original graph \cite{yu2020graph}. The GIB objective is as follows:
\begin{equation}
\begin{aligned}
\min_{G_{sub}}{ -I(G_{sub}, Y) + \beta I(G_{sub}, G)}.
\end{aligned}
\label{gib-dual}
\end{equation}
The first term in Eq.~\ref{gib-dual} encourages $G_{sub}$ to be informative to the graph label $Y$. And the second term minimizes the mutual information of $G$ and $G_{sub}$ so that $G_{sub}$ only receives limited information from the input graph $G$. The subgraph found by GIB is denoted as the \textbf{IB-subgraph}:
$G_{sub}^{*}=\mathop{\arg\min}_{G_{sub}}{ -I(G_{sub}, Y) + \beta I(G_{sub}, G)}$.
The GIB objective cannot be directly optimized since the mutual information is intractable to compute. 
Hence, it first estimates the mutual information with MINE \cite{mine}, and use the estimated value as a proxy in the optimization of GIB. This bilevel process is inefficient in optimization since it is time-consuming to estimate the mutual information. Meanwhile, inaccurate estimation also leads to an unstable training process and degenerated results. 

\begin{figure*}[t]
\begin{center}
\centerline{\includegraphics[width=2.0\columnwidth]{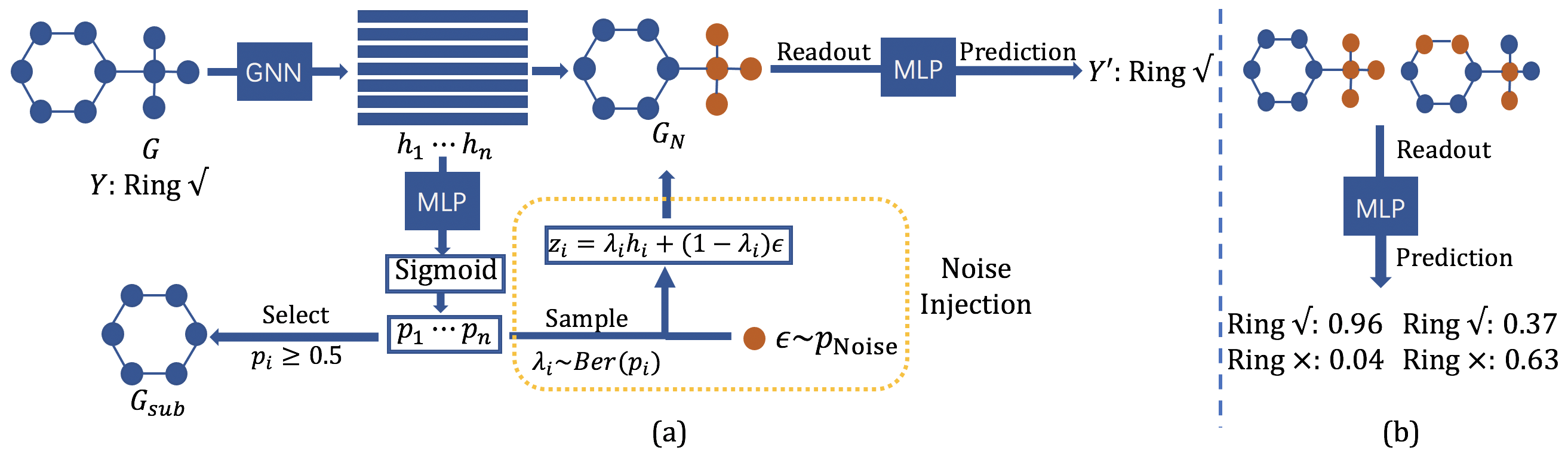}}
\end{center}
\vspace{-1.2cm}
\caption{(a) Illustration of the proposed Variational Graph Information Bottleneck (VGIB) framework. VGIB employs the noise injection method to generate a perturbed graph $G_{N}$, which is used as a "bottleneck" to distill the actionable information for predicting the graph label. The objective of VGIB has a tractable upper bound that is easy to optimize. (b) Intuition of the noise injection method. perturbing a significant subgraph is more harmful to the graph label than perturbing a label-irrelevant subgraph.}
\label{flowchart}
\vspace{-0.8cm}
\end{figure*}

\section{Method}

\subsection{Compression via Noise Injection}
Rather than directly evaluate the compression quality of $G_{sub}$ with $I(G,G_{sub})$, we inject noise into the node representations of $G$ for an alternative as shown in Fig.~\ref{flowchart}.

For an input graph $G\in \mathbb{G}$ with node feature matrix $X$, adjacent matrix $A$ and degree matrix $D$, we first generate the node representations with a $l$-layer GNN:
\begin{equation}
\begin{aligned}
H &= \mathrm{GNN}(A,X;W^{1},\cdots,W^{l})\\
 &=\underbrace{\sigma (D^{-\frac{1}{2}}AD^{-\frac{1}{2}}\cdots \sigma (D^{-\frac{1}{2}}AD^{-\frac{1}{2}}}_{l-\mathrm{layer}}XW^{1})\cdots W^{l})
\end{aligned}
\label{node-feature}
\end{equation}
where $H=[h_{1},h_{2},\cdots,h_{n}]^{T}$ is the node representation matrix. $W^{1},\cdots,W^{l}$ are the parameters at different layers.

Then we damp the information in $G$ by injecting noises into node representations with a learned probability. Let $\epsilon$ be the noise sampled from a parametric noise distribution. We assign each node a probability of being replaced by $\epsilon$. Specifically, for the i-th node,
we learn the probability $p_{i}$ with a Multi-layer Perceptron (MLP). Then, we add a Sigmoid function on the output of MLP to ensure $p_{i}\in [0,1]$:
\begin{equation}
\begin{aligned}
p_{i} = \mathrm{Sigmoid}(\mathrm{MLP}(h_{i})).
\end{aligned}
\label{probability}
\end{equation}
We then replace the node representation $h_{i}$ by $\epsilon$ with probability $p_{i}$:
\begin{equation}
\begin{aligned}
z_{i} = \lambda_{i}h_{i} + (1-\lambda_{i})\epsilon,
\end{aligned}
\label{noise-injection}
\end{equation}
where $\lambda_{i} \sim \mathrm{Bernoulli}(p_{i})$.  The transmission probability $p_{i}$ controls the information sent from $h_{i}$ to $z_{i}$. If $p_{i}=1$, then all the information in $h_{i}$ are transfered to $z_{i}$ without loss. On the contrary, when $p_{i}=0$, then $z_{i}$ contains no information from $h_{i}$ but only noise. Compared with dropping nodes for compression in GIB, this method allows to flexibly adjust the amount of information from $h_{i}$ to $z_{i}$ by changing $p_{i}$. We denote the $G_{N}=\{A,Z\}$ as the perturbed graph. At graph level, the transmissioin probability of each node determines the information sent from $G$ to $G_{N}$ in a similar way. Therefore, we can compress the information of $G$ into $G_{N}$ with a set of $p_{i}$.
We hope $p_{i}$ is learnable so that we can selectively preserve the information in $G_{N}$.
However, $\lambda_{i}$ is a discrete random variable and we can not directly calculate the gradient of $p_{i}$. Therefore, we employ the concrete relaxation \cite{jang2016categorical,gal2017concrete} for $\lambda_{i}$:

\begin{equation}
\begin{aligned}
\hat{\lambda}_{i} = \mathrm{Sigmoid}(\frac{1}{t}\log{\frac{p_{i}}{1-p_{i}}}+\log{\frac{u}{1-u}}),
\end{aligned}
\end{equation}
where $t$ is the temperature parameter and $u\sim \mathrm{Uniform}(0,1)$. Another key component in the noise injection is the specification of the injected noise.  Notice the arbitrary noise is detrimental to the semantic of the input graph. This will lead the prediction of the perturbed graph to be different from the graph property. Moreover, appropriately selected noise will endow the whole objective with a variational upper bound. Please refer to Section~\ref{assumption} for detailed discussions.



\subsection{Variational Graph Information Bottleneck}
The perturbed graph $G_{N}$ is employed to distill the actionable information of $G$ for predicting its label $Y$. 
Specifically, we compress the information of $G$ via noise injection to obtain $G_{N}$. Meanwhile, we hope that $G_{N}$ is maximally informative to $Y$, which leads to a novel Variational Graph Information Bottleneck (VGIB) framework:
\begin{equation}
\begin{aligned}
\min \limits_{G_{N}} -I(G_{N},Y)+\beta I(G_{N},G).
\end{aligned}
\label{vgib}
\end{equation}
The first term encourages $G_{N}$ to be sufficient for predicting the graph label $Y$, and the second term constrains the information that $G_{N}$ receives from $G$. These two terms require us to inject noise into $G$ selectively so that $G_{N}$ receives actionable information as much as possible. The intuition is that injecting noises into the IB-subgraph of $G$ is more harmful to the functionality of $G$ than that into the label-irrelevant substructures. In that sense, the nodes in the IB-subgraph are less likely to be injected with noise. Therefore, we can select the IB-subgraph from $G_{N}$ by this criterion after training VGIB.  We introduce the following lemma before justifying the above formulation.

\begin{lemma}
Let $G\in \mathbb{G}$ and $Y\in \mathbb{R}$ be the graph and its label. $G_{n}\in \mathbb{G}$ is the label-irrelevant substructure, which is independent to $Y$. Denote $G_{sub}$ as the arbitrary subgraph. Suppose $G_{n}$ influences $G_{sub}$ only through $G$, the following inequality holds:
\begin{equation}
\begin{aligned}
I(G_{sub},G_{n})\leq I(G_{sub},G)-I(G_{sub},Y),
\end{aligned}
\label{lemma-ineq}
\end{equation}
\label{lemma}
\end{lemma}
Lemma~\ref{lemma} indicates when setting $\beta=1$ in Eq.~\ref{gib-dual}, the GIB objective upper bounds the mutual information of $G_{sub}$ and $G_{n}$. That is to say, optimizing the GIB objective encourages $G_{sub}$ to be less related to the label-irrelevant substructure $G_{n}$. $I(G_{n},G_{sub})$ is minimal when $G_{sub}$ is the IB-subgraph.
The proof of Lemma~\ref{lemma} is in the Supplementary Materials. We next give the theorem that minimization of VGIB objective in Eq.~\ref{vgib} also leads to the irrelevance of $G_{sub}$ and $G_{n}$.

\begin{theorem}
Let $G\in \mathbb{G}$ and $Y\in \mathbb{R}$ be the graph and its label. $G_{n}\in \mathbb{G}$ is the label-irrelevant substructure in $G$, which is independent to $Y$. Denote $G_{\epsilon}\in \mathbb{G}$ as the subgraph formed by injected noise. Then, if we choose the subgraph $G_{sub}$ by dropping $G_{\epsilon}$ in $G_{N}$, the following inequality holds:
 \begin{equation}
\begin{aligned}
 I(G_{sub},G_{n}) \leq I(G_{N},G_{n}) \leq I(G_{N},G) - I(G_{N},Y)
\end{aligned}
\end{equation}
\label{theorem1}
\end{theorem}

Please refer to the Supplementary Materials for the proof of Theorem~\ref{theorem1}. VGIB differs from GIB mainly in noise injection. In the next section, we will show that this process leads to convenience in optimization. 

\subsection{Optimization of Variational Graph Information Bottleneck}
\label{assumption}

\begin{figure*}[t]
\begin{center}
\centerline{\includegraphics[width=2.0\columnwidth]{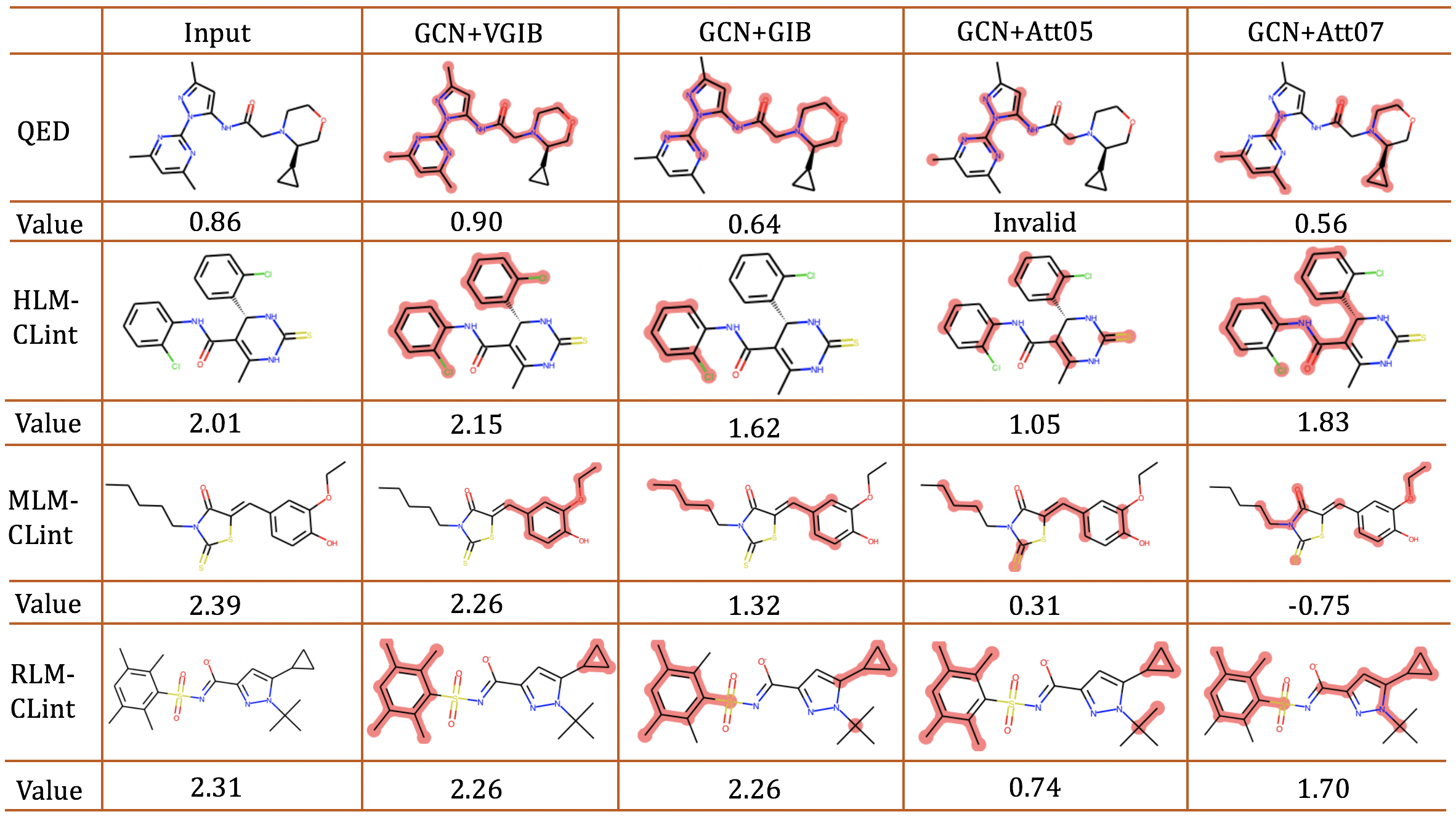}}
\end{center}
\vspace{-1.0cm}
\caption{Qualitative results on graph interpretation. VGIB can generate more precise interpretations of the input molecules.}
\vspace{-0.5cm}
\label{mi_graph}
\end{figure*}

We first examine the first term $I(G_{N},Y)$ in Eq.~\ref{vgib}, which encourages $G_{N}$ is informative of graph label $Y$. 
\begin{equation}
\begin{aligned}
-I(G_{N},Y)&\leq \mathrm{E}_{Y,G_{N}} -\log{q_{\theta}(Y|G_{N})}\\&:= \mathcal{L}_{\mathrm{cls}}(G_{N},Y),
\end{aligned}
\label{upperbound-1}
\end{equation}
where $q_{\theta}(Y|G_{N})$ is the variational approximation to $p(Y|G_{N})$. $q_{\theta}(Y|G_{N})$ outputs the label distribution of $G_{N}$ and can be modeled as a classifier. $\mathcal{L}_{\mathrm{cls}}$ is the classification loss. We choose the cross-entropy loss and mean square loss for categorical $Y$ and continuous $Y$, respectively.

For the second term $I(G_{N},G)$ in Eq.~\ref{vgib}, we first obtain the graph representation $z_{N}$ of $G_{N}$ via a readout function. We employ the sufficient encoder assumption \cite{tian2020makes} that the information of $z_{N}$ is lossless in the encoding process, leading to $I(z_{N},G) \approx I(G_{N},G)$. By choosing the distribution of noise and the readout function as the Gaussian distribution, $I(G_{N},G)$ has a tractable variational upper bound.

\begin{proposition}[Variational upper bound of $I(G_{N},G)$]
Let $m_{G}$ be the number of nodes in $G$. $h_{j}$ is the j-th node representation of $G$. $\epsilon_{G}\sim \mathcal{N}(\mu_{h},\sigma^{2}_{h})$ is the noise sampled from the Gaussian distribution. $\mu_{h},\sigma^{2}_{h}$ are mean and variance of $h_{j}$ in $G$. Suppose the readout function is chosen from mean or sum. Then up to a constant, the variational upper bound of $I(G_{N},G)$ is:
\begin{equation}
\begin{aligned}
\mathcal{L}_{\mathrm{MI}}(Z_{N},G)\leq \mathrm{E}_{G}(-\frac{1}{2}\log{A_{G}}+\frac{1}{2m_{G}}A_{G}+\frac{1}{2m_{G}}B_{G}^{2})
\end{aligned}
\label{upperbound-2}
\end{equation}
\label{prop1}
\end{proposition}
where $A_{G}=\sum_{j=1}^{m_{G}}(1-\lambda_{j})^{2}$ and $B_{G}=\frac{\sum_{j=1}^{m_{G}}\lambda_{j}(h_{j}-\mu_{h})}{\sigma_{h}}$. Please refer to the Supplementary Materials for the proof of Proposition~\ref{prop1}. One can efficiently estimate Eq.~\ref{upperbound-1} and Eq.~\ref{upperbound-2} with the batched data in the training set. The overall loss is:
\begin{equation}
\begin{aligned}
\mathcal{L} = \mathcal{L}_{\mathrm{cls}}(G_{N},Y) + \beta \mathcal{L}_{\mathrm{MI}}(Z_{N},G)
\end{aligned}
\label{loss}
\end{equation}
\begin{table}[t]
  \centering
  \caption{Mean and standard deviation of absolute property divergence between the input molecules and the found subgraphs. The results of the baselines are copied from the existing literature~\cite{yu2020graph}. The lower the better.}
  \footnotesize
  \setlength{\tabcolsep}{1.3mm}{
    \begin{tabular}{ccccc}
	\toprule
     Method     & QED   & HLM-CLint   & MLM-CLint  & RLM-CLint\\
	\midrule
    GCN+Att05 & 0.48$\pm$0.07 &  0.90$\pm$0.89 & 0.92$\pm$0.61 & 1.17$\pm$0.63 \\
    GCN+Att07 & 0.41$\pm$0.07 &  1.18$\pm$0.60 & 1.69$\pm$0.88 & 1.22$\pm$0.85 \\
    GCN+GIB & 0.38$\pm$0.12  & 0.37$\pm$0.30 & 0.72$\pm$0.55 & 1.15$\pm$0.68 \\
    GCN+VGIB & \textbf{0.32$\pm$0.12}  & \textbf{0.34$\pm$0.28} & \textbf{0.69$\pm$0.58} & \textbf{1.02$\pm$0.64} \\
	\bottomrule
    \end{tabular}}%
  \label{tab_1}%
\end{table}%

\begin{table}[t]
  \centering
  \caption{Training time of different methods on QED dataset.}
  \footnotesize
  \setlength{\tabcolsep}{1.3mm}{
    \begin{tabular}{ccccc}
    \toprule
    Method & GCN+Att05 & GCN+Att07 & GCN+GIB   & GCN+VGIB \\
    \midrule
    Time  & 142.01s & 141.18s & 1712.93s & 146.53s \\
    \bottomrule
    \end{tabular}}%
  \label{tab_efficiency}%
  \vspace{-0.5cm}
\end{table}%
\section{Experiments}
\label{experiment}
We extensively evaluate the proposed method on three tasks, i.e., graph interpretation, post-hoc explanation of GNN and graph classification. For the first task, we aim to verify whether VGIB can recognize the substructures that interpret the properties of the input molecules \cite{yu2020graph} or not. Secondly, we employ VGIB to generate post-hoc explanation of the GNNs. For the third task, we plug VGIB into various GNN baselines to see whether the found IB-subgraph can boost the performance of graph classification.
\begin{table*}[htbp]
  \centering
  \vspace{-0.5cm}
  \caption{Performance of different methods on explaianing the predictions of GCN in terms of fidelity scores. We set the sparsity score of the explanatory subgraphs as 0.5 for a fair comparison.}
    \begin{tabular}{c|cccc|cccc}
    \toprule
    Metric & \multicolumn{4}{c|}{Fidelity+$\uparrow$} & \multicolumn{4}{c}{Fidelity-$\downarrow$} \\
    \midrule
    Property  & \multicolumn{1}{c}{RLM-Clint} & HLM-Clint & QED   & DRD2  & RLM-Clint & HLM-Clint & QED   & DRD2 \\
    \midrule
    GNNExplainer & 0.694 & 0.778 & 0.602 & 0.74  & 0.478 & 0.616 & 0.498 & 0.433 \\
    PGExplainer & 0.632 & 0.692 & 0.598 & 0.686 & 0.502 & 0.62  & 0.56  & 0.54 \\
    GraphMask & 0.632 & 0.706 & 0.602 & 0.673 & 0.516 & 0.592 & 0.574 & 0.4866 \\
    IGExplainer & 0.684 & 0.758 & 0.592 & 0.693 & 0.602 & 0.686 & 0.584 & 0.58 \\
    GraphGrad-CAM & 0.67  & 0.782 & 0.586 & 0.659 & 0.56  & 0.668 & 0.564 & 0.566 \\
    GIB   &  0.654    & 0.781 & 0.601 & 0.724 & 0.483 & 0.643 & 0.525 & 0.428 \\
    VGIB  & \textbf{0.765} & \textbf{0.792} & \textbf{0.627} & \textbf{0.756} & \textbf{0.463} & \textbf{0.579} & \textbf{0.487} & \textbf{0.424} \\
    \bottomrule
    \end{tabular}%
  \label{tab-fidelity}%
  \vspace{-0.5cm}
\end{table*}%
\subsection{Graph Interpretation}
In this experiment, we extract the substructures which have the most similar properties to the original molecules. We consider four properties: QED, HLM-CLint, MLM-CLint, and RLM-CLint. QED measures the probability of a molecule being a drug within the range of \([0, 1.0]\). HLM-CLint, MLM-CLint, and RLM-CLint are estimated values of in vitro human, mouse and rat liver microsome metabolic stability, respectively (base  10 logarithm of mL/min/g)\footnote{We evaluate QED values of molecules with the toolkit on \url{https://www.rdkit.org/}. Moreover, we obtain HLM-CLint, MLM-CLint, and RLM-CLint value of molecules on \url{https://drug.ai.tencent.com/}.}. We collect molecules with QED$\geq 0.85$, HLM-CLint$\geq 2$, MLM-CLint$\geq 2$, and RLM-CLint$\geq 2$ from ZINC250K \cite{irwin2005zinc}, and we individually build datasets with the selected molecules for the four properties. For each property, we use $85\%$, $5\%$, and $10\%$ of the molecules for training, validating, and testing, respectively. Please refer to the Supplementary Materials for the statistics of datasets.


We compare the proposed VGIB with GIB \cite{yu2020graph} and the attention-based method \cite{hierG}. For VGIB, we first learn the node representation with a GCN and inject noise into each node as shown in Eq.~\ref{noise-injection} and Eq.~\ref{probability}. Then we obtain the perturbed graph representation by pooling the noisy node representations with a readout function. We thereafter optimize the loss function in Eq.~\ref{loss} and collect the nodes with $p_{i}\geq 0.5$ in Eq.~\ref{probability} to obtain the IB-subgraph. We simultaneously supervise the classifier in Eq.~\ref{upperbound-1} with the representation of the input graph  and its label to enhance the informativeness of the subgraph. For GIB, we follow the existing method~\cite{yu2020graph} and optimize the model in a bilevel optimization scheme. As for the attention-based method, we attentively pool the node representations with the attention scores for label prediction. Furthermore, we select the nodes with top $50\%$ and $70\%$ attention scores to form the predictive subgraphs. For each method, we adopt a 2-layer GCN with 16 hidden dimensions for a fair comparison. We run experiments on one TITAN RTX GPU. If the found subgraph is disconnected, we choose its largest connected part to ensure chemical validity.

We report the mean and standard deviation of absolute property divergence between the input molecules and the found subgraphs in Table~\ref{tab_1}. We quantitatively compare the performance of different methods on the interpretation of graph properties. The proposed VGIB method shows favorable performance against GCN+GIB. 
For the attention-based method, the performance is sensitive to the selected value of the threshold since the results of GCN+Att05 vary from those of GCN+Att07. Therefore, one needs to finetune the threshold for different tasks. In contrast, our VGIB is free from a manually selected threshold thanks to the information-theoretic objective.

We then compare the training time of different methods on the QED dataset. We train different methods 3 times and report the average training time in Fig.~\ref{tab_efficiency}. It is shown that the attention-based methods achieve the fastest training due to their simple architectures. Although VGIB is slower than the attention-based methods, it achieves significant performance gain in finding subgraphs with more similar properties to the input molecules. Compared with GIB, VGIB trains over ten times faster than GIB with better performance. The reason is that estimating the mutual information in GIB is time-consuming. However, VGIB a enjoys tractable upper bound of its objective, which is easy to optimize.

\begin{table*}[t]
  \centering
  \caption{Classification accuracy of graphs with \textbf{sum pooling} and \textbf{mean pooling} as the readout function. We report the mean and standard deviation of the testing accuracy in 10-fold cross-validation for each method. The bold results are the overall best performances and the underlining results are the best performances of certain backbones.}
	\setlength{\tabcolsep}{0.7mm}{
    \begin{tabular}{c|ccccccc}
	\toprule
    \multicolumn{1}{c}{}&Method & MUTAG & PROTEINS & IMDB-B & DD    & COLLAB & REDDIT-B \\
	\midrule
    \multirow{12}{*}{\rotatebox{90}{Sum pooling}} &GCN   & 0.76$\pm$0.09&	0.72$\pm$0.05&	0.71$\pm$0.04&	0.74$\pm$0.03&	0.78$\pm$0.03 &0.75$\pm$0.05 \\
    &GIB+GCN & 0.77$\pm$0.07&\underline{0.74$\pm$0.04}&0.72$\pm$0.04 & 0.75$\pm$0.05&	0.78$\pm$0.02&	0.77$\pm$0.04 \\
    &VGIB+GCN & \underline{0.79$\pm$0.09} &	\underline{0.74$\pm$0.04}&	\underline{0.74$\pm$0.04}&	\underline{0.77$\pm$0.09}&	\textbf{0.80$\pm$0.02}&	\underline{0.82$\pm$0.02} \\
	\cmidrule{2-8}
    &GraphSAGE & 0.74$\pm$0.08&	0.73$\pm$0.05&	0.71$\pm$0.05&	0.75$\pm$0.03&	0.78$\pm$0.02	&0.78$\pm$0.10 \\
    &GIB+GraphSAGE & 0.75 $\pm$0.07&	0.73$\pm$0.04&	0.72$\pm$0.05&	0.76$\pm$0.04&	0.79$\pm$0.02&	\underline{0.80$\pm$0.03} \\
    &VGIB+GraphSAGE & \underline{0.77$\pm$0.07}&	\underline{0.74$\pm$0.04}&	\underline{0.73$\pm$0.03}&	\textbf{0.78$\pm$0.05}&	\textbf{0.80$\pm$0.03}&	\underline{0.80$\pm$0.09} \\
	\cmidrule{2-8}
    &GIN   & 0.83$\pm$0.07&	0.74$\pm$0.05&	0.71$\pm$0.05&	0.71$\pm$0.03&	0.78$\pm$0.02	&0.81$\pm$0.10 \\
    &GIB+GIN & 0.84$\pm$0.06&	0.74$\pm$0.05&	0.74$\pm$0.07	& 0.74$\pm$0.04&	0.78$\pm$0.03&	0.84$\pm$0.03 \\
    &VGIB+GIN & \textbf{0.86$\pm$0.08}&	\textbf{0.75$\pm$0.04}	&\textbf{0.74$\pm$0.05}	&\underline{0.75$\pm$0.04}&	\underline{0.79$\pm$0.03}	&\textbf{0.86$\pm$0.03} \\
	\cmidrule{2-8}
	    &GAT   & 0.75$\pm$0.08&	0.72$\pm$0.04&	0.72$\pm$0.03&	0.75$\pm$0.04& 0.76$\pm$0.04& 0.71$\pm$0.08 \\
    &GIB+GAT & 0.75$\pm$0.09&	\underline{0.73$\pm$0.04}	&0.72$\pm$0.05&	0.76$\pm$0.04&	0.77$\pm$0.03&	0.73$\pm$0.04 \\
    &VGIB+GAT & \underline{0.76$\pm$0.07}&	\underline{0.73$\pm$0.04}	&\underline{0.73$\pm$0.07}	&\underline{0.77$\pm$0.04}	&\underline{0.79$\pm$0.03}	&\underline{0.76$\pm$0.05} \\
	\midrule
	\midrule
    \multirow{12}{*}{\rotatebox{90}{Mean pooling}}&GCN   & 0.72$\pm$0.11&	0.71$\pm$0.04	&0.71$\pm$0.04&	0.72$\pm$0.05&	0.77$\pm$0.02&	0.75$\pm$0.06 \\
    &GIB+GCN & 0.74$\pm$0.08&	0.72$\pm$0.05	&0.72$\pm$0.03&	0.74$\pm$0.05&	0.78$\pm$0.02&	0.76$\pm$0.04 \\
    &VGIB+GCN & \underline{0.76$\pm$0.10}&	\underline{0.73$\pm$0.04}&	\underline{0.73$\pm$0.04}&	\underline{0.75$\pm$ 0.10}&	\textbf{0.79$\pm$0.02}&	\underline{0.78$\pm$ 0.02} \\
	\cmidrule{2-8}
    &GraphSAGE & 0.73$\pm$0.07&	0.72$\pm$0.04	&0.70$\pm$0.04	&0.73$\pm$0.04&	0.77$\pm$0.02&	0.78$\pm$0.03 \\
    &GIB+GraphSAGE & 0.74$\pm$0.07&	0.72$\pm$0.04	&0.72$\pm$0.05&	0.74$\pm$0.04	& \underline{0.78$\pm$0.03}&	0.79$\pm$0.03 \\
    &VGIB+GraphSAGE & \underline{0.75$\pm$0.08}&	\underline{0.73$\pm$0.04}&	\underline{0.73$\pm$0.03}&	\textbf{0.76$\pm$0.05}&	\underline{0.78$\pm$0.03}&	\underline{0.80$\pm$0.03} \\
	\cmidrule{2-8}
    &GIN   & 0.82$\pm$0.07&	0.71$\pm$0.06&	0.72$\pm$0.05&	0.73$\pm$0.03&	0.78$\pm$0.03&	0.82$\pm$0.02 \\
    &GIB+GIN & 0.83$\pm$0.06&	0.71$\pm$0.05&	0.73$\pm$0.07&	0.74$\pm$0.04&	0.78$\pm$0.03&	0.84$\pm$0.04 \\
    &VGIB+GIN & \textbf{0.84$\pm$0.09}&	\underline{0.72$\pm$0.05}&	\textbf{0.74$\pm$0.05}&	\underline{0.75$\pm$0.04}&	\underline{0.79$\pm$0.03}	& \textbf{0.85$\pm$0.03} \\
	\cmidrule{2-8}
    &GAT   & 0.74$\pm$0.08&	0.71$\pm$0.04&	0.71$\pm$0.04&	0.70$\pm$0.06&	0.76$\pm$0.03&	0.77$\pm$0.04 \\
    &GIB+GAT & 0.75$\pm$0.10&	0.73$\pm$0.04	&0.72$\pm$0.05&	0.72$\pm$0.04&	0.77$\pm$0.03&	\underline{0.79$\pm$0.04} \\
    &VGIB+GAT & \underline{0.76$\pm$0.07}&	\textbf{0.74$\pm$0.03}&	\underline{0.73$\pm$0.07}&	\underline{0.73$\pm$0.05}&	\underline{0.78$\pm$0.03}&	\underline{0.79$\pm$0.05} \\
	\bottomrule
    \end{tabular}}%
  \label{tab:sum_mean}%
  \vspace{-0.6cm}
\end{table*}%

\subsection{Explainability of GCN}
As GCN model becomes prevalent in various tasks \cite{conf/nips/HamiltonYL17,yu2022structure}, there is increasing concerns on its explainability since GCN are treated as black-box \cite{gnnexplainer}.
In this section, we employ the proposed VGIB to explain the prediction of GCN for molecule classification on ZINC250K dataset. We consider four properties: QED, HLM-CLint, RLM-CLint and DRD2 construct the dataset for each property. For QED, we label the molecules with QED$\geq$0.85 /$<$0.85 as 1/0. For DRD2, we label the molecules with DRD2$\geq$0.50 /$<$0.50 as 1/0. For HLM-CLint and RLM-CLint, we label the molecules with the property value greater than 2.0 as 1 or 0 otherwise. For each property, we train the GCN on the training set and employ VGIB to generate post-hot explanation on the test set. Please refer to Supplementary Materials for more details on the dataset splits and the training process. 

We compare VGIB with various explanation model including GNNExplainer \cite{gnnexplainer}, PGExplainer \cite{luo2020parameterized}, GraphMask \cite{schlichtkrull2020interpreting}, IGExplainer \cite{sundararajan2017axiomatic}, GraphGrad-CAM \cite{pope2019explainability} and GIB \cite{yu2020graph}. These methods interpret the prediction of GCN with the node importance score ranging in $[0,1]$, where 1 indicates the node is the most important to the molecule classification and 0 otherwise. Please refer to Supplementary Materials for more details on the baseline methods. 

We employ the fidelity score to evaluate how the explanation is faithful to the GCN model \cite{yuan2021explainability}. Specifically, let $y_{i}$ and $\hat{y}_{i}$ be the ground-truth and the prediction of the i-th input molecule. Define $k$ as the sparsity score of the explanatory subgraph. The explanatory subgraph is obtained by choosing the nodes with top $k$\% scores from the input molecule, and we denote its prediction as $\hat{y}_{i}^{k}$. The Fidelity- score is computed as follows:
\begin{equation}
\begin{aligned}
Fidelity-=\frac{1}{N}\sum_{i=1}^{N}\mathds{1}(y_{i}=\hat{y}_{i})-\mathds{1}(y_{i}-\hat{y}_{i}^{k})
\end{aligned}
\end{equation}
where $\mathds{1}(y_{i}=\hat{y}_{i})$ is the indicator function which outputs 1 if $y_{i}=\hat{y}_{i}$ and 0 otherwise. Fidelity- score measures how the prediction of the explanatory subgraph is close to the input molecule. The lower value of Fidelity- score indicates more faithful explanation. Similarly, we define $\hat{y}_{i}^{1-k}$ as the prediction of the complementary subgraph, which is obtained by removing the explanatory subgraph from the input. The Fidelity+ score is defined as follows:
\begin{equation}
\begin{aligned}
Fidelity+=\frac{1}{N}\sum_{i=1}^{N}\mathds{1}(y_{i}=\hat{y}_{i})-\mathds{1}(y_{i}-\hat{y}_{i}^{1-k})
\end{aligned}
\end{equation}
The Fidelity+ score indicates more important nodes are identified in the explanation. 

\begin{figure*}[t]
\begin{center}
\centerline{\includegraphics[width=2.0\columnwidth]{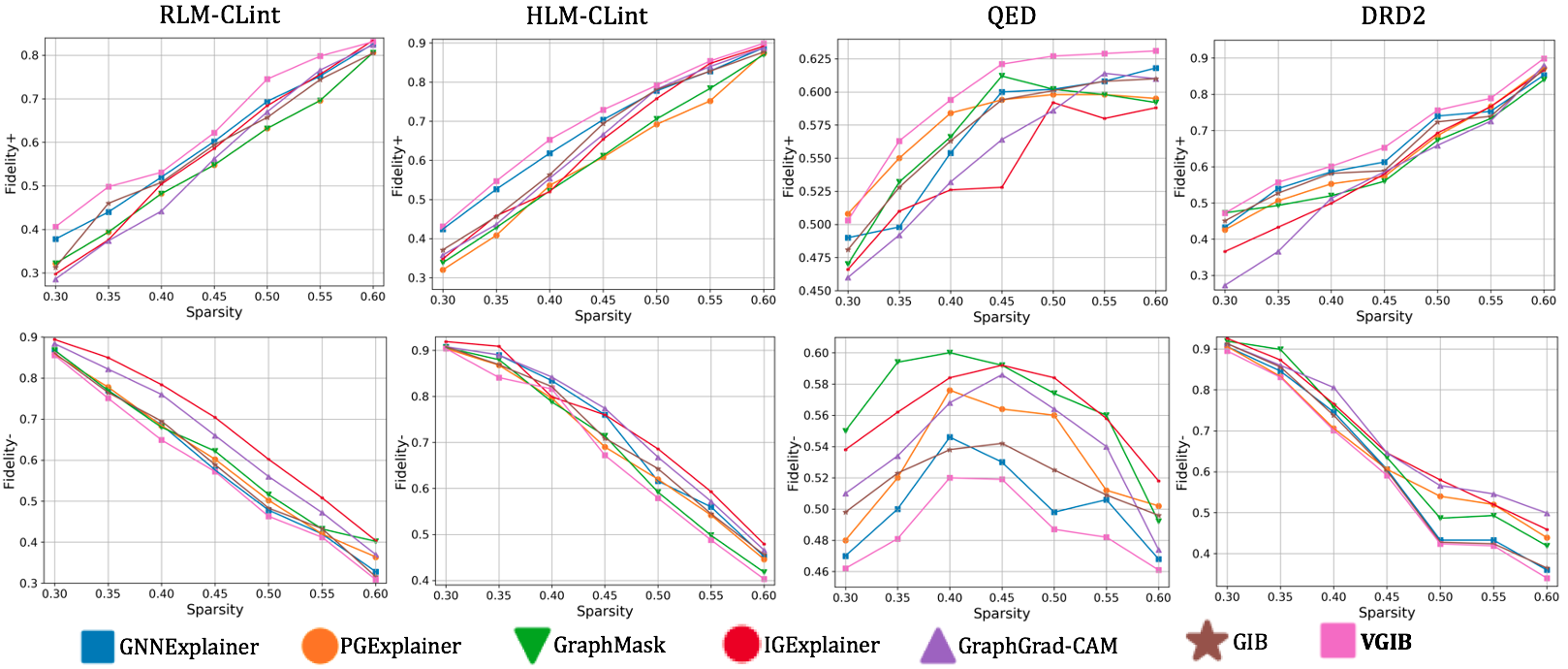}}
\end{center}
\vspace{-0.8cm}
\caption{Performance of different methods on explaining the predictions of GCN in terms of fidelity scores ($Fidelity+\uparrow$ and $Fidelity-\downarrow$). VGIB generates the most faithful explanations to the predictions of GCN by identifying the most important nodes under different sparsity scores.}
\label{ibplane}
\vspace{-0.6cm}
\end{figure*}

Table~\ref{tab-fidelity} shows the fidelity scores of the explanations produced by different methods. The sparsity score is set to be $k=0.5$ for all the explanatory subgraphs. As shown in table~\ref{tab-fidelity}, VGIB achieves best fidelity scores on all the properties. This shows that VGIB generate faithful explanation to the predictions of the GCN. Moreover, to comprehensively evaluate different methods, we set $k\in\{0.30,0.35,0.40,0.45,0.50,0.55,0.60\}$ and compare the performance under different sparsity scores. As shown in Figure~\ref{ibplane}, VGIB generate most faithful explanations to the predictions of GCN in terms of fidelity scores by under different sparsity scores. This shows that VGIB can identify most important nodes to the predictions of GCN.

\subsection{Graph Classification}
In this subsection, we aim to find out whether the found subgraph can improve the performance of baselines on graph classification or not. We evaluate different methods on MUTAG \cite{rupp2012fast}, PROTEINS \cite{borgwardt2005protein}, DD, IMDB-BINARY, REDDIT-BINARY, and COLLAB \cite{nr} datasets, which are wildly used for graph classification. Please refer to Supplementary Materials for the statistics of datasets.

We consider four GNN baselines including GCN~\cite{gcn}, GAT ~\cite{velickovic2017graph}, GraphSAGE ~\cite{conf/nips/HamiltonYL17} and GIN ~\cite{Xu:2019ty}. 
We use the mean and sum pooling as a readout function for the baseline methods to obtain the graph representation for prediction. Then, similar to GIB \cite{yu2020graph}, we plug VGIB into these baselines. Specifically, we adopt the baseline models to extract node representation. Then, we recognize the IB-subgraph by optimizing the VGIB objective. We pool the node representations in the IB-subgraph for classification with the same readout function as the baselines. For a fair comparison, we adopt a 2-layer network architecture and 16 hidden dimensions for different methods. We train these methods for 100 epochs and test the models with the smallest validation loss. We report the mean and standard deviation of accuracy across 10 folds \footnote{We follow the protocol in \url{https://github.com/rusty1s/pytorch_geometric/tree/master/benchmark/kernel}}.

The experimental results are summarized in Table~\ref{tab:sum_mean}. Compared with the baselines, VGIB can discover an informative yet compressed subgraph of the input. Therefore, it relieves the perturbation of noise structures and redundant information and boosts the performance of the baselines. VGIB also outperforms the GIB-based methods on most of the datasets.

\section{Discussions}
\textbf{Potential Negative Impacts:} Our method aims to discover a predictive yet compressed subgraph of the input graph, and can be deployed in social network analysis and biochemistry. The concern is that if not adequately used under administration, our method potentially leads to the leakage in privacy and intellectual property.

\textbf{Limitations:} We assume the encoding process from $G_{N}$ to $z_{N}$ is lossless following the sufficient encoder assumption \cite{tian2020makes}. Hence we approximate the compression term $I(G_{N},G)$ with $I(z_{N},G)$ and obtain a tractable variational upper bound for optimization. In fact, the encoding process is not lossless due to the data processing inequality. Thus, we actually approach $I(G_{N},G)$ with $I(z_{N},G)$ in practice. We leave the in-depth analysis in our future work.

\section{Conclusion}

%
%
We propose a novel Variational Graph Information Bottleneck framework for improved and efficient subgraph recognition.
The proposed noise injection method serves as an alternative to compress the information in the discovered subgraph and allows a tractable objective of VGIB for efficient and stable training. 
Using the proposed method, we make the practical training more than 10 times faster than the existing methods.
The experimental results show that the proposed method performs favorably against the existing methods on various tasks.

\section*{Acknowledgement}
This work is partially funded by National Natural Science Foundation of China (Grant No. U21B2045, U20A20223) and Youth Innovation Promotion Association CAS (Grant No. Y201929).

{\small
\bibliographystyle{ieee_fullname}
\bibliography{egbib}
}

\end{document}